\documentclass[pdflatex,sn-mathphys]{sn-jnl}

\usepackage{makecell}
\usepackage{amssymb}
\usepackage{bm}
\usepackage{float}
\usepackage{hyphsubst}
\usepackage{makecell}
\HyphSubstLet{english}{usenglishmax}
\usepackage[english]{babel}
\usepackage{rotating}

\usepackage{soul}
\soulregister\ref{7}
\soulregister\cite{7}

\theoremstyle{thmstyleone}%
\theoremstyle{thmstyletwo}%

\theoremstyle{thmstylethree}%

\begin{document}

\title[Domain Adaptive Sim-to-Real Segmentation of Oropharyngeal Organs]{Domain Adaptive Sim-to-Real Segmentation of Oropharyngeal Organs}


\author[1]{\fnm{Guankun} \sur{Wang}}\email{gkwang@link.cuhk.edu.hk}

\author[2,3]{\fnm{Tian-Ao} \sur{Ren}}\email{taren@buct.edu.cn}

\author[1]{\fnm{Jiewen} \sur{Lai}}\email{jiewen.lai@cuhk.edu.hk}
\author[1]{\fnm{Long} \sur{Bai}}\email{b.long@link.cuhk.edu.hk}
\author*[1]{\fnm{Hongliang} \sur{Ren}}\email{hlren@ee.cuhk.edu.hk; ren@nus.edu.sg}

\affil[1]{\orgdiv{Department of Electronic Engineering}, \orgname{The Chinese University of Hong Kong}, \state{Hong Kong}, \country{China}}

\affil[2]{\orgdiv{College of Mechanical and Electrical Engineering}, \orgname{Beijing University of Chemical Technology}, \state{Beijing}, \country{China}}

\affil[3]{\orgdiv{Shenzhen Research Institute}, \orgname{The Chinese University of Hong Kong}, \city{Shenzhen}, \state{Guangdong}, \country{China}}

\abstract{Video-assisted transoral tracheal intubation (TI) necessitates using an endoscope that helps the physician insert a tracheal tube into the glottis instead of the esophagus. The growing trend of robotic-assisted TI would require a medical robot to distinguish anatomical features like an experienced physician which can be imitated by utilizing supervised deep-learning techniques. However, the real datasets of oropharyngeal organs are often inaccessible due to limited open-source data and patient privacy. In this work, we propose a domain adaptive Sim-to-Real framework called \textbf{I}oU-\textbf{R}anking \textbf{B}lend-\textbf{A}rt\textbf{F}low (IRB-AF) for image segmentation of oropharyngeal organs. The framework includes an image blending strategy called IoU-Ranking Blend (IRB) and style-transfer method ArtFlow. Here, IRB alleviates the problem of poor segmentation performance caused by significant datasets domain differences; while ArtFlow is introduced to reduce the discrepancies between datasets further. A virtual oropharynx image dataset generated by the SOFA framework is used as the learning subject for semantic segmentation to deal with the limited availability of actual endoscopic images. We adapted IRB-AF with the state-of-the-art domain adaptive segmentation models. The results demonstrate the superior performance of our approach in further improving the segmentation accuracy and training stability.}

\keywords{Domain adaption, Semantic segmentation, Sim-to-Real Transfer}



\maketitle

\section{Introduction}\label{sec1}

Transoral tracheal intubation (TI) is the gold standard for securing a patient's airway when they require respiratory assistance. However, the success of this procedure hinges on the physician's skills to correctly insert an endotracheal tube into the patient's trachea \cite{thomas2014tracheal}. Asphyxia, hypoxia, and pulmonary aspiration can cause severe morbidity and death if the TI is not performed in a timely manner \cite{caplan2003practice}. With advances in robotics and AI technology, image-guided automation that uses visual information to plan, complete, and recognize specific tasks is becoming an emerging area in medical robotics \cite{lu2021toward, lai2021variable, lu2022unified} and rehabilitation \cite{bruce2022mmnet}. On top of video-assisted transoral TI, robot-assisted transoral TI makes intubation even more effective through automation. Yet, robotization requires the robot to distinguish and understand the contour of the anatomical features like an experienced physician. Therefore, it is expected that the robot should segment the endoscopic vision with sufficient fidelity. The above-mentioned initiative motivates the work herewith.

Segmenting oropharyngeal organs is one of the most critical steps in robot-assisted intubation. In practice, semantic segmentation divides each pixel into a label, and each label is assigned to a class, which can be applied to radiation, image-guided therapies, and enhanced radiological diagnostics \cite{asgari2021deep}. However, obtaining real datasets of oropharyngeal organs is challenging due to patient privacy and the need for sufficient open-source data. Deploying supervised learning techniques for segmentation tasks with insufficient data would result in poor performance of the segmentation network model. To alleviate this problem, one can exploit synthetic or simulation data for Sim-to-Real transfer \cite{frangi2018simulation}. The method generally trains the network using rich simulation data while the tests are performed in real data that lacks abundance. Current simulation frameworks in the medical field, such as Blender \cite{rehman2022organs}, SOFA \cite{duriez2013control}, and so on, provide users with increasingly realistic 3D scenarios with more sensations like textures and interaction-responsive physical properties. All anatomical structures can be rotated, scaled, and moved at any angle, and the hierarchy and adjacency of organs can be viewed from the inside out. They not only aid in providing a potentially abundant data source \cite{zhao2020sim} but also relieve privacy concerns with real datasets.

Nonetheless, there are some challenges in employing virtual data for Sim-to-Real in deep learning. For instance, the open-source project Surgical Blender can build virtual organ textures and morphologies based on advanced computer graphic techniques. Still, the pre-processing of this framework can be dramatically complex and time-consuming, which leads to limited accessibility of the data \cite{ganry2017use, pierri2019bimaxillary}. While Surgical Blender is designed explicitly for virtual surgical simulation, SOFA is a relatively mature and high-performance library for general-purpose physical simulations. Although SOFA can easily handle general 3D models with a broader range of accessibility, their textures and morphologies are usually missing. In this work, we choose SOFA-generated images as the source domain, since data generation is more accessible with higher efficiency. For the problem of lower-quality data, if our work can improve the segment performance of models in Sim-to-Real, a better demonstration can be expected for domain adaption of other general data. The model trained with simulation images may have difficulty in segmentation compared with the model trained with real images, as the oropharyngeal organs in simulation images are constrained in representing natural textures like color and reflection. Deep learning models that are sensitive to data have yet to guarantee an effective generalization to unrestricted test cases. Therefore, segmentation performance will be greatly degraded in real datasets (i.e., target domains) if the model training relies on simulated datasets only, making the Sim-to-Real deployment challenging. 

\begin{figure}
    \centering
    \includegraphics[width=4in]{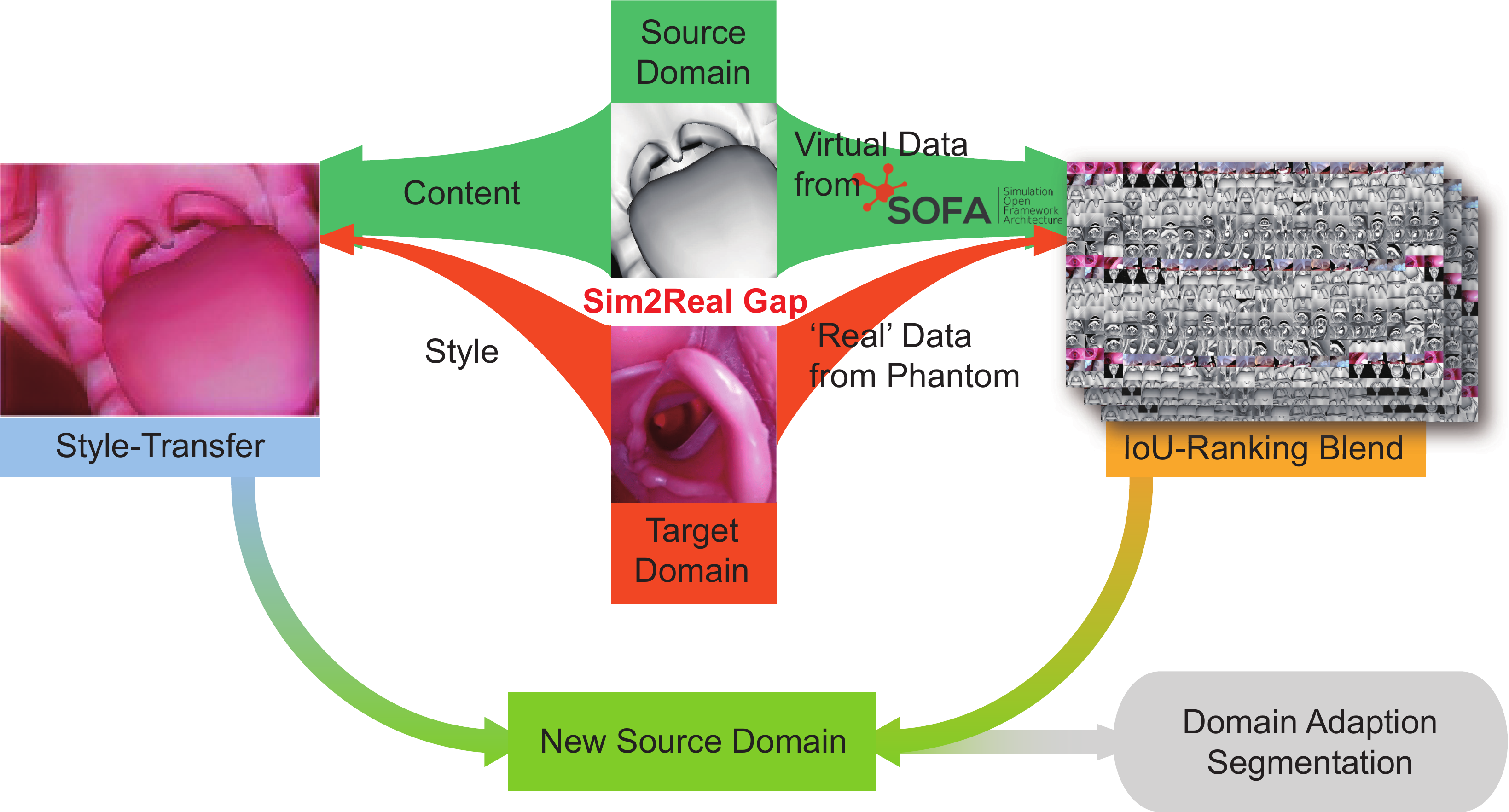}
    \caption{Overview of the proposed IRB-AF framework. The style-transfer module first narrows the difference (i.e., the sim-to-real gap) between two domains. Then the source domain is reconstructed based on IoU-Ranking Blend. The new source domain will be used as the training set for segmentation.}
    \label{Figure1}
\end{figure}

Given the abovementioned problems, reducing the differences between datasets is the most direct and effective way. Among all, domain randomization and unsupervised domain adaptation are often used. Domain randomization, as a commonly used technique in Sim-to-Real transfer learning, is often used to increase the robustness of a model to real-world variability \cite{chen2021understanding}. It is generally done by randomizing various aspects of the simulation environment, such as the texture, lighting, and even physical property of objects, followed by model training on the randomized simulation data to adapt to different environment variations \cite{tobin2017domain}. The training process shall be repeated until the model is sufficiently robust after being evaluated by real-world data.

Unsupervised Domain Adaptation (UDA) refers to the process of adapting a model that has been trained on annotated samples from one distribution (the source domain) to function on another distribution (the target domain) for which no annotations are provided \cite{yang2020fda}. Most recent UDA works focus on lessening the difference between the two datasets during training on the source domain. Maximum Mean Disparity (MMD) and its kernel variants are proposed by \cite{geng2011daml,long2015learning} as a common measure of discrepancy. Central Moment Discrepancy (CMD) \cite{zellinger2017central} extended the measure to higher-order statistics. In addition, some researchers used pseudo labels provided by self-training \cite{zou2018unsupervised}, or target data \cite{wu2018dcan, sankaranarayanan2018learning} produced by generative networks. Traditional domain adaptation methods \cite{yang2020fda, vu2019advent, hoffman2018cycada} demonstrate excellent domain adaptation performances in the datasets with minor differences like GTA5 \cite{richter2016playing}, SYNTHIA \cite{ros2016synthia}, and CityScapes \cite{cordts2016cityscapes}. However, the model's performance will decline rapidly when the differences become more significant.

Therefore, semi-supervised learning (SSL) techniques are required. During SSL, the model is adapted to the dataset with one of its subsets annotated. By properly aligning domains, SSL can become unsupervised domain adaptation. There are some common methods between SSL and UDA, such as self-training \cite{li2019bidirectional, zou2018unsupervised}, class balancing \cite{zhu2005semi}, and generative model. As one of the successful approaches, entropy minimization is also used in semi-supervised learning \cite{springenberg2015unsupervised}. Both the adversarial loss of the entropy maps and the entropy of the pixel-wise prediction are minimized by \cite{vu2019advent}.  

In this work, we use the SOFA framework to reconstruct a virtual scene where a steerable robotic endoscope navigates through a 3D oropharynx model with its endoscopic vision recorded. Besides, real images are captured on a real-world phantom. To improve the performance of UDA models, we propose a domain adaptive Sim-to-Real framework called \textbf{I}oU-\textbf{R}anking \textbf{B}lend-\textbf{A}rt\textbf{F}low (IRB-AF) that includes a novel image-blending strategy and the style-transfer method. Firstly, we blend a small batch of real images into the simulation domain with the Intersection over Union (IoU)-Ranking Blend (IRB) mechanism. The mechanism sorts the resultant IoU among classes after training and refines the blending proportion for the next training iteration according to the sorting. In this regard, the potential of a limited number of mixed images can be fully utilized in the training process to improve the segmentation performance of real domains. Then, the style-transfer technique ArtFlow \cite{an2021artflow} is used to reduce the differences between the source and target domains. In practice, IRB-AF combines these two methods by first modifying the style of the source domain images and then blending target domain images via IRB. The overview of this framework is depicted in Fig.~\ref{Figure1}.


This work contributes the following:
\begin{itemize}
    \item An image segmentation method targeting oropharyngeal organs with domain adaptive Sim-to-Real transfer;
    \item A novel IRB approach aiming at reducing the domain gap between virtual and real datasets for segmentation accuracy improvement;
    \item A style-transfer-based domain adaptive segmentation strategy to improve the network's training stability.
    \item An open-sourced dataset of endoscopic images generated from SOFA-based oropharynx model with style transfer from phantom (EISOST).
\end{itemize}

The rest of this paper is organized as follows. Section \ref{SOFA-gen} introduces the data preparation for the image segmentation task. Section \ref{domain adaptive} illustrates the domain adaptive sim-to-real convention based on the proposed IRB-AF strategy. Section \ref{experiment} showcases the performance of the IRB-AF strategy based on our dataset. The final section concludes the paper.

\section{SOFA-Generated Virtual Dataset}\label{SOFA-gen}
\begin{figure}[t!]
    \centering
    \includegraphics[width=4in]{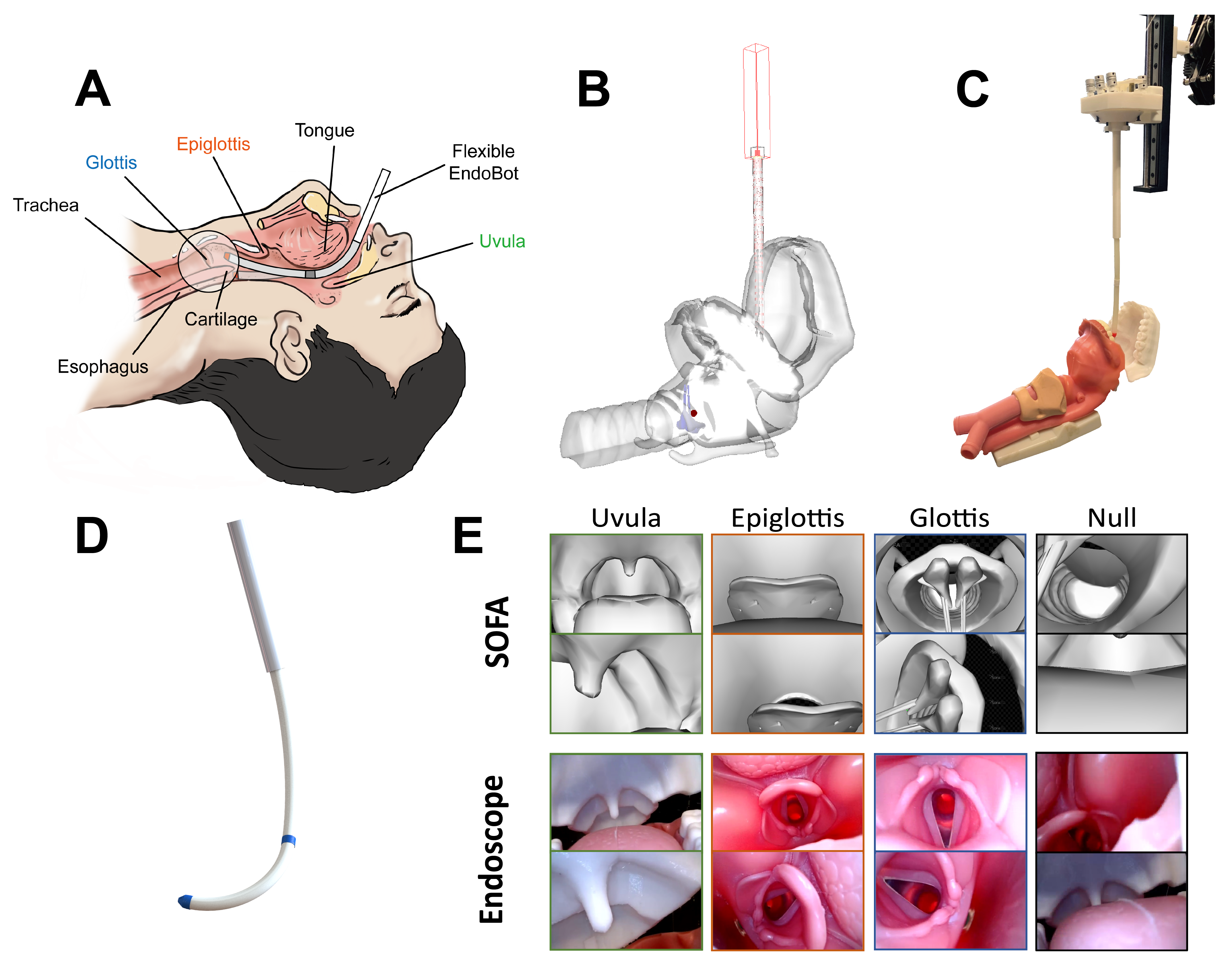}
    \caption{(A) Using a flexible robotic endoscope (EndoBot) as a stylet to guide the next step's tracheal intubation. (B) SOFA scene. (C) Real-world scene. (D) CAD design of our EndoBot (adapted from \cite{lai2022constrained}) that was used in the virtual and reality setup. (E) Dataset examples. ``Null'' indicates there is no target in the scene.}
    \label{Figure2}
\end{figure}

As shown in Fig. \ref{Figure2}A, robot-assisted TI employs a steerable flexible endoscope that works as a stylet to navigate to the repository tract instead of the digestive tract with the aid of endoscopic vision. The automation of such a process requires the robot to recognize or even segment the oropharyngeal organs by learning from a dataset with a great number of medical images. However, real-world endoscopic images of the oropharynx are often inaccessible due to ethical and privacy issues. While using different phantoms for image collection can be tedious and time-consuming, we propose to employ virtual endoscopic images as the dataset to train the  segmentation network. In the virtual environment, one can conveniently generate an abundance of datasets \cite{zhao2020sim} of different types for deep learning. Several virtual environments are capable of establishing 3D anatomical scenes, such as Blender \cite{rehman2022organs}, SOFA \cite{duriez2013control}, etc. In previous work \cite{unpublished}, we built an interactive environment of a soft robotic endoscope and oropharynx in the simulation (SOFA v22.06.99) as shown in Fig. \ref{Figure2}B and \ref{Figure2}D. The rich SOFA image data also motivated us to develop a virtual dataset-based segmentation approach, which could be useful in real-world soft robotic applications. However, the over-animated scenes pose a major challenge to the Sim-to-Real transfer.

\subsection{Generation of Virtual Data}

In recent years, researchers have been trying to develop large-scale virtual datasets to supplement the limited real datasets. For instance, \cite{allan2019endovis17, allan2020endovis18} release datasets and challenges to evaluate the state-of-the-art in surgical image segmentation and improve surgeon capabilities. With excellent virtual engines and rendering, modern computer games can also be used to generate virtual data sets to replace realistic scenes that require high costs to obtain \cite{richter2016playing}. In \cite{unpublished}, a 3D oropharyngeal phantom modified based on \cite{3dmodel} was imported into the virtual scene. The phantom includes three necessary oropharyngeal organs, namely, the uvula, epiglottis, and glottis. To reduce the expensive finite element computation, we trimmed the insignificant entities from the phantom, such as teeth and miscellaneous muscles. Using virtual images allows one to generate a large dataset with customized differentiation like the field of view, color, texture, and model variety, which significantly reduces the dependence on limited real images.

\subsection{Preparation of Training Data}
The training dataset is composed of two parts. The first part consists of the virtual images obtained by a flexible endoscopic robot navigating to the desired spot near the glottis in SOFA, as shown in Fig. \ref{Figure2}B. The second part includes real images obtained by a similar setup and manner in the phantom environment in the real world, as shown in Fig. \ref{Figure2}C. Some sample images are demonstrated in Fig. \ref{Figure2}E. The size of the initial dataset labeled with bounding boxes is given in Table \ref{tab1}. {As some images contain multiple organs, the available images in virtual and real sets are 1194 and 203, respectively.} For the annotations, we provide coarse and fine annotations at the pixel level, including instance-level labels for oropharyngeal organs.
\begin{table}[t!]
	\renewcommand{\arraystretch}{1}
	\caption{Size of the Blended Dataset for Oropharyngeal Organ Recognition\\{(Unit: Frame)}}
	\label{tab1}
	\centering
	\begin{tabular}{cccc}
		\hline
		& Uvula & Epiglottis & Glottis\\
		\hline
		SOFA's Virtual Images & 601 & 395 & 507\\
		Colored Real Images & 119 & 69 & 82\\
		\hline
	\end{tabular}
\end{table}

\section{Domain Adaptive Sim-to-Real with IRB-AF}
\label{domain adaptive}

In this section, we introduce our domain adaption segmentation in two aspects. The proposed IRB approach is a compelling dataset blending strategy used for the Sim-to-Real training, while the image style-transfer is used to further reduce the differences between the source domain and the target domain, thereby improving the domain adaption performances.  {We integrate the above two methods and propose IRB-AF that aligns the image distributions of different datasets in terms of content and style.}

\subsection{IRB: IoU-Ranking Blend}

Conventional domain adaption methods focus on modifying the image features  {(e.g., color features, texture features, shape features, and spatial relationship features of an image)} to reduce the distribution between the source and target domains. However, it is noticeable that only minor image differences occur in their dataset selection in the first place. This is often seen in applications like urban scenes and city traffic images. Yet, in the medical image field, obtaining virtual images that are significantly recognizable (i.e., being photo-realistic) to the real organs or anatomical features would be impractical. To deal with this problem, we propose to mix a small batch of images from the target domain during the training whenever using virtual or synthetic medical images, { which can increase the content similarity between datasets.} Since the distribution of features differs among images, the segmentation performance based on randomly-mixed training will be unstable. To maximize the usability and potential of a designated small batch of blended images, we propose the IRB based on the model training performance. The IoU-Ranking Blend method mixes images based on each class's segmentation testing results after training the model.

Since our dataset (see Table \ref{tab1}) contains three classes (excluding the background/null), we devise the following criteria to formulate the blending:
\begin{enumerate}
    \item The number of blending images is a multiple of 10 to facilitate statistics and differentiate the outcomes of blending;
    \item The proportion of blends is increased among classes to ensure lower-ranked classes have larger weights; 
    \item The least number of classes is a multiple of 2 in case the segmentation performance drops rapidly due to too insufficient proportions.
\end{enumerate}

\begin{figure*}[t!]
    \centering
    \includegraphics[width=4in]{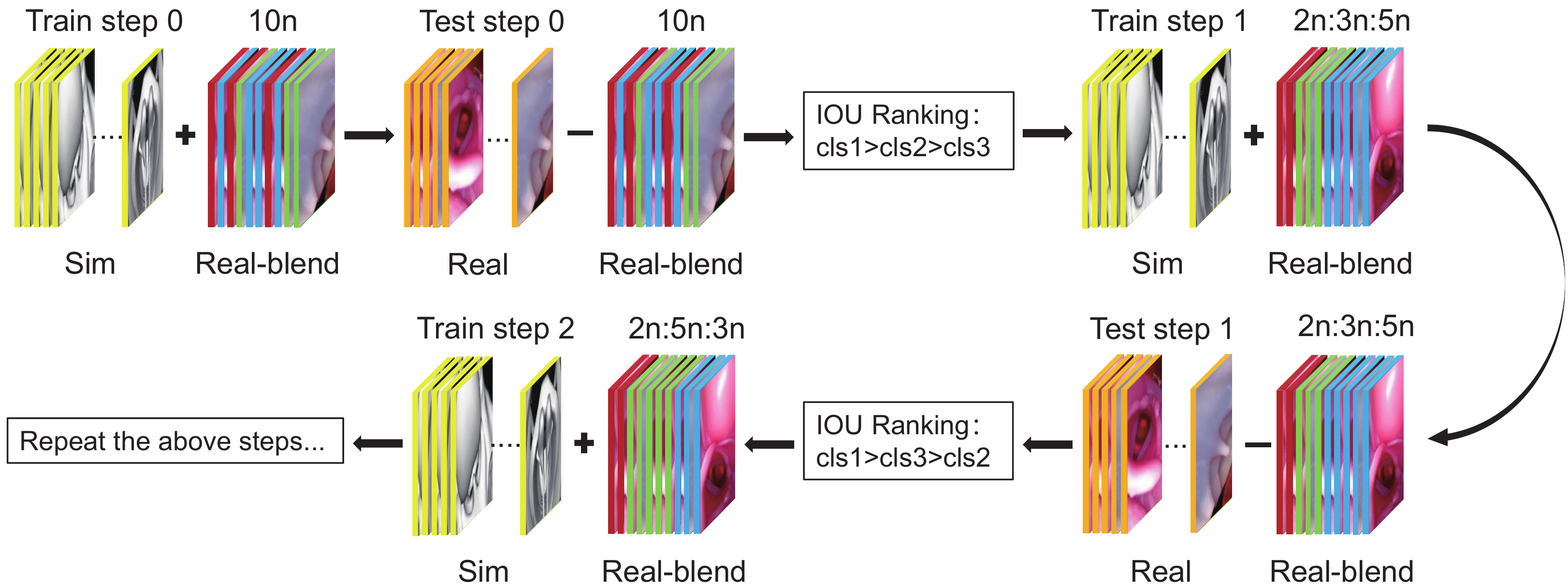}
    \caption{Flow chart of IoU-Ranking Blend. \emph{10n} and \emph{cls} represent the number of blended images in the class, respectively. The colors of three \emph{cls} are red, green, and blue. The above steps will be repeated until there is no new IoU ranking. The best mIoU will be selected from all test results.}
    \label{Figure3}
\end{figure*}

Therefore, a ratio of 5:3:2 for each class that fulfills the above criteria was chosen to investigate the IRB strategy, while the number of blends is 10 to 40. The resultant percentage of blending images in each subset is shown in Table \ref{tab2}. A maximum blending of 40 real images would ensure a `virtual purity' of over 90\% for each organ. Fig. \ref{Figure3} shows the flow chart of the blending sequence. 
\begin{table}[t!]
	\renewcommand{\arraystretch}{1}
	\caption{Percentage of blending images in each subset}
	\label{tab2}
	\centering
	\begin{tabular}{cccc}
		\hline
		\# of Blends & Uvula & Epiglottis & Glottis\\
		\hline
		10 & 1.637\% & 2.469\% & 1.934\%\\
		20 & 3.221\% & 4.819\% & 3.795\%\\
            30 & 4.754\% & 7.059\% & 5.587\%\\
            40 & 6.240\% & 9.195\% & 7.313\%\\
		\hline
	\end{tabular}
\end{table}

In the first step of training, randomly distributed images $I_b = \{x_t\}_{i=1}^{10n}$ where $x_t \in \mathbb{R}^{H\times W\times 3}$ are mixed in source domain. When the first training is completed, the segmentation results of the model on the remaining images in the target domain reflect the model's performance for each class. During the training process, the model's sensitivity on different oropharyngeal organs reflects differently, so the number of mixed images required for different classes to achieve the same testing performance varies. For some classes, if the differences between their source and target domains are minor, or the features are sharper compared to the background, they would require a lighter blend to improve the Sim-to-Real performance. However, other classes may require more to achieve the same effect.

{After training, the resultant IoU of the test dataset is ranked in descending order. Then, the proportion of blends can be adjusted as deemed. For example, if the IoU ranking ends up with $\{\textrm{IoU}_A > \textrm{IoU}_B > \textrm{IoU}_C\}$ where $\{A, B, C\}$ represents the classes, it indicates that class $C$ demonstrates relatively poor performance in the segmentation and the blending proportion of class $C$ needs to be increased. In contrast, the blending proportion of class $A$ can be reduced appropriately due to its better performance.} {For the next step's training set, the dataset is proportionally partitioned as $I_b = \{\{x_t^C\}_{i=1}^{5n},\{x_t^B\}_{i=1}^{3n}, \{x_t^A\}_{i=1}^{2n}\}$. To avoid the model learning features outside the blending images, we randomly initialize the model parameters at the beginning of each training step. Although the performance of some classes could slightly reduce in the testing after adjusting the blending ratios, the rest could perform better, thus improving the training's mean of IoU (mIoU). Moreover, with the continuous adjustment of each training iteration, the ranking of IoUs will be gradually fixed. After that, we can opt for their optimal ratio so that the potential for a limited amount of mixed images can be exploited to the greatest extent. The experiments in the next section will demonstrate the effectiveness of our strategy by comparing random and optimal proportions.}

\subsection{Image Style-Transfer for Domain Adaption}

The style-transfer provides a new viewpoint to narrow the differences between the source and target domain in terms of image style.  {Here, the style essentially refers to the textures, colors, and visual patterns in images at various spatial scales, and they are considered low-level image features.} {A typical style-transfer strategy can be described as optimizing the weighted sum of content loss $\mathcal{L}_c$ and style loss $\mathcal{L}_s$~\cite{ghiasi2017exploring}, which can be defined as follows:}
\begin{equation}
\mathcal{L}_s=\left\|\mathcal{G}\left[\mathcal{A}(t)\right]-\mathcal{G}\left[\mathcal{A}(s)\right]\right\|_F^2
\end{equation}
\begin{equation}
\mathcal{L}_c=\left\|\mathcal{A}(t)-\mathcal{A}(c)\right\|_2^2.
\end{equation}
Here, {$\left\|...\right\|_F^2$ and $\left\|...\right\|_2^2$ are the Frobenius Norm and the Euclidean Norm, respectively. \textit{s} and \textit{c} indicate lower layers and higher layers in an image classification network.} $\mathcal{A}(t)$ is the network activations, and $\mathcal{G}(t)$ denotes the Gram matrix of the network activations. The variables $s$ and $c$ represent the lower and higher layers, respectively.
Neural style-transfer (NST) aims to transfer the style of the target domain into the image of the source domain to enhance the similarity of the features from different domains. However, due to the information loss caused by pooling~\cite{li2017universal}, the training bias from the loss function~\cite{huang2017arbitrary}, and the biased style-transfer module~\cite{liao2017visual}, the content leak problem occurs---the content information of the source domain may be lost during the style-transfer training process~\cite{an2021artflow}. Therefore, we employ ArtFlow~\cite{an2021artflow}, which includes an unbiased feature transfer module and reversible neural flows to prevent content leaks during the style-transfer. ArtFlow establishes the Projection Flow Network (PFN) following the Glow model~\cite{kingma2018glow} and replaces the traditional encoder-decoder structure with a projection-reversion strategy. The components of PFN (additive coupling~\cite{dinh2014nice}, invertible 1$\times$1 convolution~\cite{kingma2018glow}, and Actnorm~\cite{kingma2018glow}) are completely reversible, which also ensures that information is lossless when transmitted through PFN.

\begin{figure}[t!]
    \centering
    \includegraphics[width=4in]{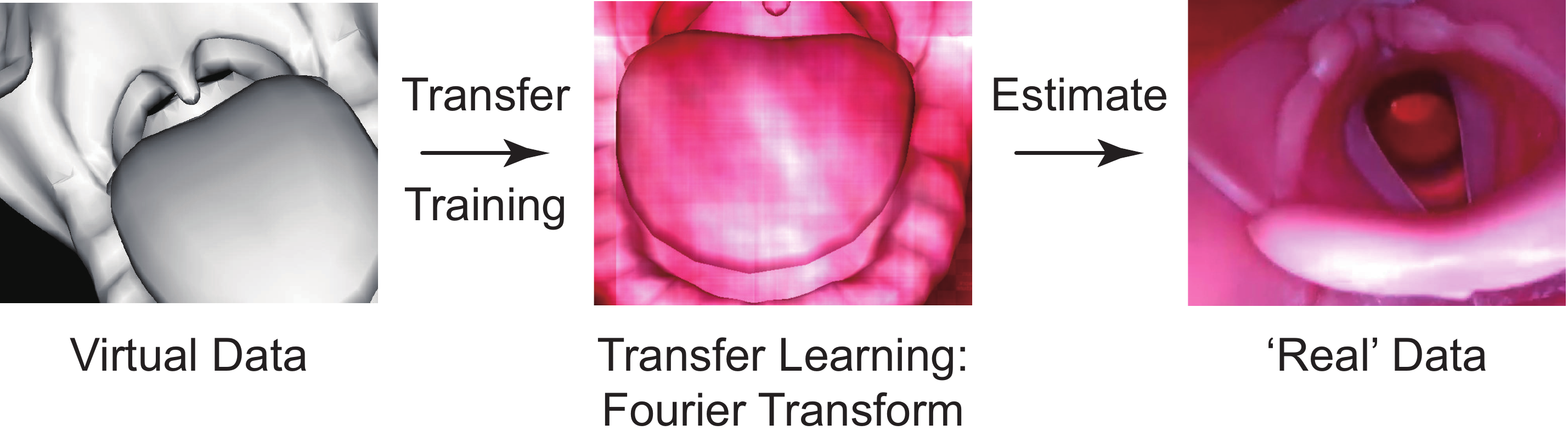}
    \caption{{Reducing the sim-to-real gap between the virtual and `real' data using transfer learning based on Fourier Transform. There are still appearance gaps between virtual images and real images.}}
    \label{Figure4}
\end{figure}

\begin{figure}[t!]
    \centering
    \includegraphics[width=4in]{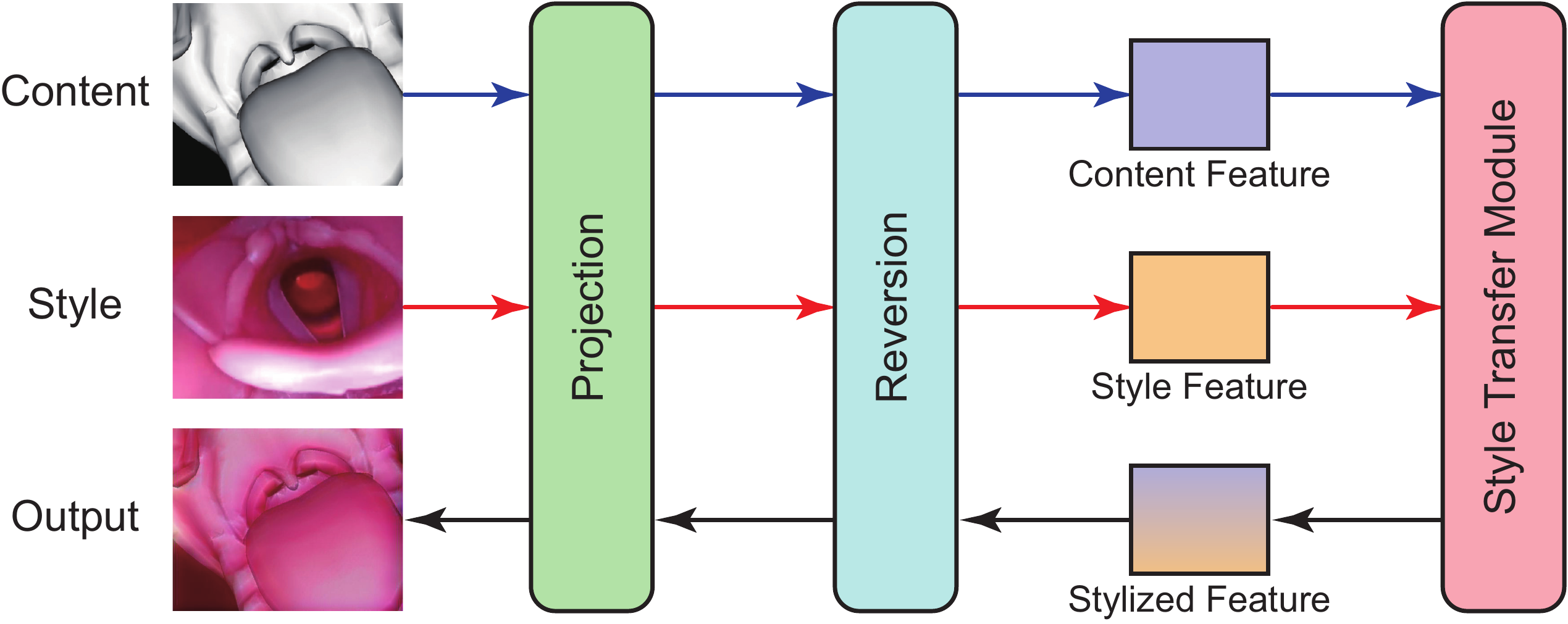}
    \caption{{The overview of employing ArtFlow in this work. Instead of the auto-encoder-based framework, ArtFlow applies a projection-transfer-reversion scheme. The projection module extracts content and style images' features which will be used to generate stylized features in Transfer Module. Afterward, the stylized feature is transferred to the stylized image via reversion inference.}}
    \label{Figure5}
\end{figure}

We have made numerous efforts to generate synthetic images. However, they still have a certain appearance gap with real images and cannot be directly used to train oropharyngeal organ segmentation. An example is shown in Fig.~\ref{Figure4}. With the help of ArtFlow, we try to convert the appearance of virtual images into real oropharyngeal organs' appearance, thereby enhancing the sense of photo-realistic of virtual data while preserving useful anatomical features for model training. A schematic of employing style-transfer to the virtual data is shown in Fig.~\ref{Figure5}.  {Image style-transfer helps us to reduce the differences between datasets from the low-level features. And the image content with high-level features is optimized by the IoU-Ranking Blend method in the previous subsection. In practice, IRB-AF first modifies the style of the source domain images and then blends a small number of images via IRB, which jointly modifies the distribution between the source domain and target domain, and improves the performance of the segmentation model from the perspective of the low-level and high-level features of the images, respectively.}
{Therefore, our dataset contains three parts, i.e., endoscopic images generated from SOFA-based oropharynx model, endoscopic images captured from real-world phantom, and style-transferred virtual images. We have published this dataset under the name EISOST\footnote{\href{https://github.com/gkw0010/EISOST-Sim2Real-Dataset-Release}{Endoscopic Images generated from SOFA-based oropharynx model with style transfer from phantom (EISOST) - https://github.com/gkw0010/EISOST-Sim2Real-Dataset-Release}}.}

\section{Experiments} \label{experiment}

 {To demonstrate the effectiveness of the proposed IRB-AF on domain adaptation segmentation, we conduct extensive experiments. We make a comparison before and after the introduction of IRB-AF on the state-of-the-art domain adaptive segmentation models. Moreover, an ablation experiment is made to qualitatively analyze the respective effects of IoU-Ranking Blend and style-transfer in domain adaptation segmentation.}

\begin{sidewaystable}
\sidewaystablefn%
\footnotesize
\begin{center}
\begin{minipage}{\textheight}
\caption{Performances of Different SOTA Segmentation Methods Adopting our IRB-AF Strategy}
\label{tab3}
\begin{tabular*}{\textheight}
{@{\extracolsep{\fill}}cccccccccccccc@{\extracolsep{\fill}}}
\toprule  
\multirow{2}{*}{Method}	&	\multirow{2}{*}{Train Step}	&	    	&	\multicolumn{5}{c}{\makecell*[c]{Intersection over Union}} 	&			&			\multicolumn{5}{c}{\makecell*[c]{Accuracy}}					\\ \cline{4-8} \cline{10-14}
	&	    	& \multirow{11}{*}{} &	background	&	glottis	&	epiglottis	&	uvula	&	mIoU	& \multirow{11}{*}{}   	&	background	&	glottis	&	epiglottis	&	uvula	&	mAcc	\\  
\cline{1-2} \cline{4-8} \cline{10-14}
	&	40-r 	&	    	&	94.990 	&	72.030 	&	54.540 	&	65.660 	&	71.805 		&			&	97.73	&	85.19	&	59.44	&	87.34	&	82.425 	\\
FDA~\cite{yang2020fda}	&	40-253	&	    	&	96.620 	&	77.280 	&	74.260 	&	63.890 	&	78.013 		&			&	98.21	&	86.55	&	88.16	&	79.78	&	88.175 	\\
	&	40-235	&	    	&	96.490 	&	78.440 	&	73.980 	&	66.590 	&	\textbf{78.875} 		&			&	98.15	&	86.4	&	88.47	&	80.18	&	\textbf{88.300} 	\\
\cline{1-2} \cline{4-8} \cline{10-14}
	&	40-r 	&	    	&	96.147 	&	80.790 	&	65.060 	&	67.900 	&	77.474 		&			&	98.71	&	87.79	&	70.17	&	80.78	&	84.363 	\\
Advant \cite{vu2019advent} &	40-253	&	    	&	97.100 	&	84.060 	&	79.330 	&	64.790 	&	\textbf{81.320} 		&			&	99.21	&	83.33	&	92.29	&	71.85	&	86.670 	\\
	&	40-235	&	    	&	96.600 	&	78.290 	&	76.630 	&	67.460 	&	79.745 	&				&	98.69	&	82.41	&	89.35	&	79.22	&	\textbf{87.418} 	\\
\cline{1-2} \cline{4-8} \cline{10-14}
	&	40-r 	&	    	&	95.100 	&	72.930 	&	62.830 	&	59.080 	&	72.485 	&				&	97.51	&	77.95	&	89.42	&	66.41	&	82.823 	\\
Cycada \cite{hoffman2018cycada}	&	40-253	&	    	&	94.640 	&	75.320 	&	72.380 	&	64.890 	&	\textbf{76.808} 	&				&	97.02	&	87.31	&	72.6	&	76.16	&	83.273 	\\
	&	40-235	&	    	&	96.040 	&	69.790 	&	62.290 	&	63.620 	&	72.935 	&				&	97.89	&	85.36	&	87.62	&	74.88	&	\textbf{86.438} 	\\
\bottomrule 
\end{tabular*}
\end{minipage}
\end{center}
\end{sidewaystable}

\subsection{Evaluation and comparison on different networks}

\label{experiment1}
\{\emph{1) Implementation details:} We train and test three domain adaptive segmentation models, namely, FDA~\cite{yang2020fda}, ADVENT~\cite{vu2019advent} and CyCADA~\cite{hoffman2018cycada}, to validate the generality of our methods. The training is conducted on an Nvidia GeForce RTX 3090 GPU, and the batch size is set to 4 in all our experiments. We take DeepLabV2~\cite{chen2017deeplab} with ResNet101~\cite{he2016deep} as the baseline. The optimization algorithm is selected as stochastic gradient descent (SGD), and the learning rate is $5\times 10^{-3}$ with a momentum of 0.9 and weight decay of $5\times 10^{-4}$. For the training parameters of ArtFlow, we will show them in Section~\ref{ST}. The larger the number of blended images, the lower the improvement in Sim-to-Real performance, which is demonstrated in Section~\ref{experiment2}. As a result, we just experimented with 40 blending images. If there is an improvement with this setting, it is inevitable with less blending quantity. 

\emph{2) Results and discussion:} 
The results are shown in Table~\ref{tab3} in which IRB-AF achieves higher mIoU and mean Accuracy (mAcc) in terms of Sim-to-Real performance. Compared to the original settings of these models, although increasing the number of blended images weakens the performance improvement, our method still increases 4.96$\%$--9.85$\%$ in mIoU and 3.62$\%$--7.13$\%$ in mAcc, respectively.

\subsection{Ablation Experiment}
\subsubsection{IoU-Ranking Blend in FDA} 
\label{experiment2}
\emph{1) Implementation:} Fourier Domain Adaptation (FDA) is one of state-of-the-art domain adaptive segmentation models. The main idea of FDA is to use Fourier transform to swap the low-frequency spectra of the source domain and target domain images to reduce their differences. The experimental parameters are the same as those in Section~\ref{experiment1}. We bring in the IRB strategy and adapt it to the FDA. 

Since the improvement would be significantly weakened when the blending quantity exceeds 40, we set 40 as the upper limit in our quantitative experiments. Note that a blend of 40 images accounts for only 3\% of the training set. For each blended set, the same training and test will be conducted three times to reduce the randomness of training, and all records will be made. 

\begin{figure}[t!]
    \centering
    \includegraphics[width=4in]{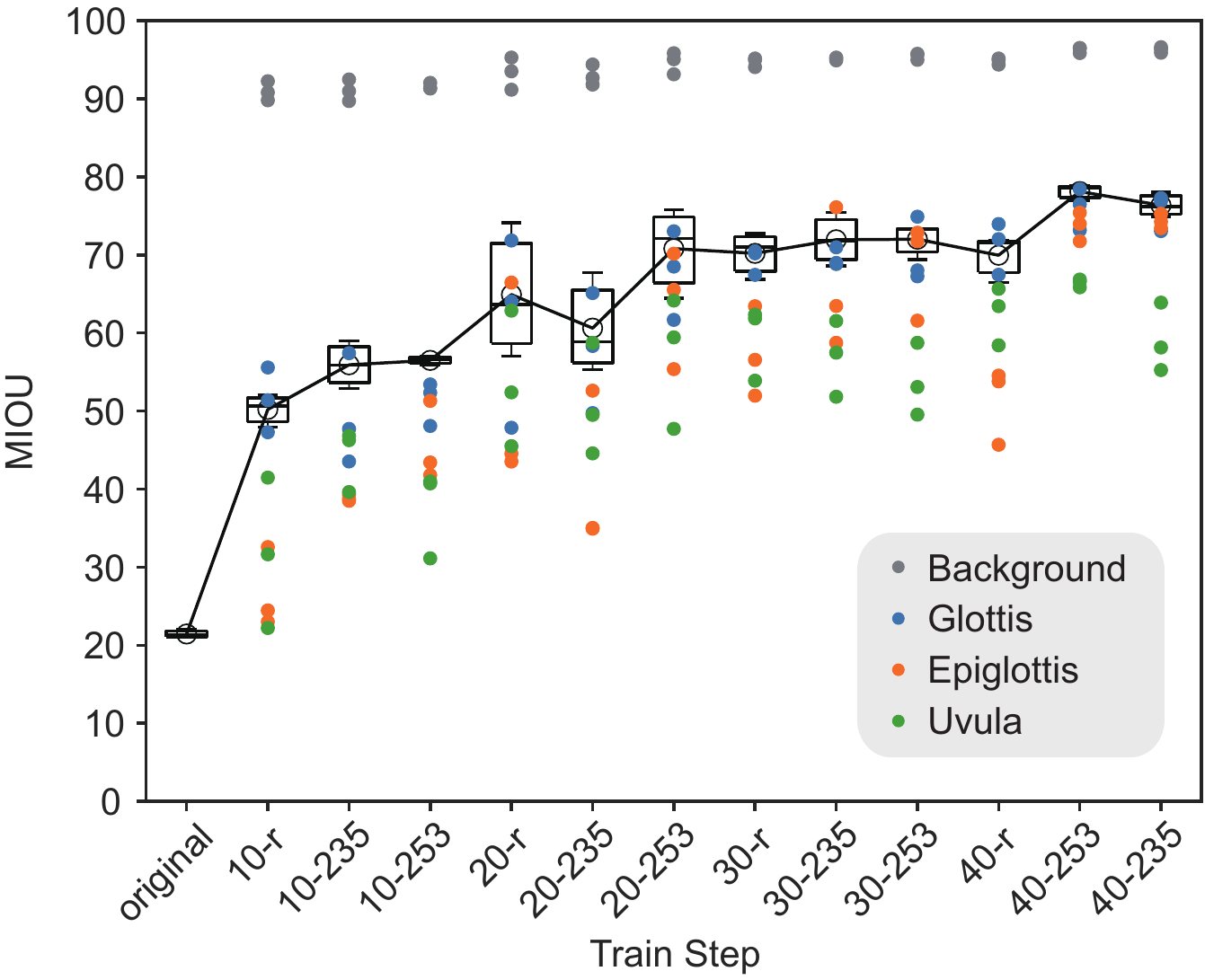}
    \caption{The mIoU comparison between the IRB-FDA and original FDA. The horizontal axis represents the training sequence. Each element of the boxplot contains the results of three experiments, which are also shown by scatter points. The data of the line chart comes from the mean of mIoU.}
    \label{Figure6}
\end{figure}

\begin{sidewaysfigure}
\centering
\includegraphics[width=7.5in]{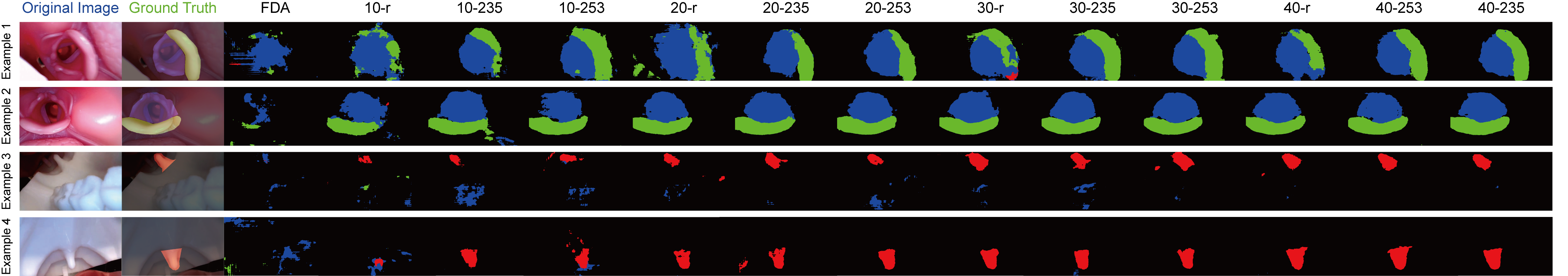}
    \caption{Performance comparison on segmenting in the real domain of different oropharyngeal organs.}
    \label{Figure7}
\end{sidewaysfigure}

\emph{2) Results and discussion:} The mIoU of different blending groups is shown in Fig. \ref{Figure6}. Due to the Sim-to-Real gap between the source and target domain, the original FDA may not achieve a satisfactory domain adaptation performance with low-frequency spectrum exchange. Consequently, the mIoU of the test set has always been low at about 21\%. However, when ten real images are randomly mixed in the training set (the ratio between the original and the blended images is about 120:1), the mIoU of multiple tests can sharply increase to over 50\%. Moreover, after introducing our IRB strategy, the model fully uses the blended images according to the classes, making the average mIoU exceed 56\% with the lower limit exceeding 55\%. When further increasing the mixed quantity, although the mIoUs fluctuate, their mean of upper and lower limits gradually ascended. It is shown that IRB can increase mIoU by more than 5\% under different blending quantities. The qualitative comparison of segmentation output is shown in Fig. \ref{Figure7}.

\begin{table}[t!]
\caption{Segmentation results on FDA Style-Transfer}
\renewcommand\arraystretch{.98}
\label{tab4}
\begin{center}
\begin{tabular}{ccccc}
\toprule  
Train Step&	mIoU& * &	Stability	& * \\
\midrule  
10-r	&	46.495 	&	↓ 3.712	&	2.048 	&	↓ 2.055	\\
10-235	&	50.861 	&	↓ 5.061	&	4.178 	&	↓ 1.915	\\
10-253	&	53.493 	&	↓ 2.998	&	5.810 	&	\textcolor{red}{↑} 4.793	\\
20-r	&	61.432 	&	↓ 3.486	&	3.480 	&	↓ 13.620	\\
20-235	&	62.694 	&	\textcolor{red}{↑} 2.070	&	5.558 	&	↓ 6.870	\\
20-253	&	72.005 	&	\textcolor{red}{↑} 1.199	&	4.313 	&	↓ 7.008	\\
30-r	&	70.543 	&	\textcolor{red}{↑} 0.340	&	2.387 	&	↓ 3.503	\\
30-235	&	73.198 	&	\textcolor{red}{↑} 2.379	&	2.758 	&	↓ 5.225	\\
30-253	&	73.143 	&	\textcolor{red}{↑} 0.683	&	3.895 	&	↓ 0.425	\\
40-r	&	72.996 	&	\textcolor{red}{↑} 3.034	&	3.373 	&	↓ 1.945	\\
40-235	&	74.723 	&	↓ 1.642	&	6.153 	&	↓ 3.013	\\
40-253	&	75.823 	&	↓ 2.296	&	0.632 	&	↓ 1.328\\
\bottomrule 
\end{tabular}
\end{center}
\textrm{*Compared to non-style-transfer.}
\end{table}

\subsubsection{Style-Transfer in FDA}
\label{ST}
 {\emph{1) Implementation details:} For the ArtFlow model, we set two blocks in the backbone network and eight flow modules in each block. The training of ArtFlow is carried out according to the \cite{an2021artflow} so that the optimization algorithm is Adam~\cite{kingma2014adam} instead of SGD, and the learning rate is $1\times10^{-4}$ with weight decay of $5\times10^{-5}$. The number of training iterations is set to 120,000. The configuration opts empirically. }

After training the ArtFlow, each source image will be assigned to a random target image with the same oropharyngeal organs as style-transfer input to ensure that the style of the target domain can be fully utilized. Then, we set the style-transfer images as the new source domain for FDA training. Note that the experimental parameters are the same as those in Section~\ref{experiment1}. 

\emph{2) Results and discussion:} 
{The experimental results are shown in Table~\ref{tab4}.} We focus on the mean of mIoU and the stability of repeated experiments with different mixed combinations to find the best training configuration for future applications. According to the results, the mean mIoU of the tests does not generally increase but increases when the blending proportion becomes larger. It shows that the improvement of style-transfer on mIoU is limited, and the contribution to the segmentation accuracy improvement of the target domain is mainly due to the blended images from the target domain. However, the data fluctuations of multiple tests of style-transfer are much lower than that of previous experiments, which shows that the style-transfer can improve the training stability and reduce the impact of randomness. The results indicate that the style-transfer could enhance the stability of semi-supervised domain adaptive segmentation training by mitigating variations between datasets, although the improvement in segmentation accuracy is limited.

\section{Conclusion}
 {In this paper, we propose a domain adaptive Sim-to-Real framework called IRB-AF. The framework includes an image blending strategy called IoU-Ranking Blend (IRB) and style-transfer method ArtFlow.} The IRB solves the accuracy degradation problem caused by the significant difference between simulation and real datasets during unsupervised domain adaptive segmentation of oropharyngeal organs. By sorting the segmentation results between classes, a more appropriate blending strategy can be formulated to maximize the potential of employing a limited number of blended images. The style-transfer method ArtFlow is introduced to reduce the differences between datasets further. The Sim-to-Real segmentation results show that our proposed method improves the performance of the existing domain adaptive segmentation models. Furthermore, we also found that the style-transfer can enhance training stability and alleviate the impact of randomness.

{Although we have made some progress in oropharyngeal organ segmentation, the accuracy of the model still has much space for improvement. Future work can be expected to enrich domain adaptation segmentation methods in the medical field. Besides, the enrichment of real datasets is also significant to further improve the accuracy of the model.}

\backmatter

\bmhead{Supplementary information}
This article has an accompanying supplementary video. 

\bmhead{Acknowledgments}
This work was supported in part by the Hong Kong Research Grants Council (RGC) Collaborative Research Fund (CRF-C4026-21GF) 

\section*{Declarations}


\begin{itemize}
\item Conflict of interest/Competing interests: No benefits in any form have been or will be received from a commercial party related directly or indirectly to the subject of this manuscript.
\item Ethics approval: Ethical approval was not sought for the present study because this article does not contain any studies with human or animal subjects.
\item Authors' contributions: G.W., J.L., and H.R. conceived the concepts. G.W., T.R., J.L., and L.B. advised on the design and implementation of the experiments. G.W., T.R., and J.L. conducted experiments and analyzed the data. G.W., T.R., J.L., and L.B. wrote the manuscript. All authors read, edited, and discussed the manuscript and agree with the claims made in this work. H.R. coordinated and supervised the research.
\end{itemize}





\bigskip









\bibliography{sn-article}


\begin{thebibliography}{45}
\ifx \bisbn   \undefined \def \bisbn  #1{ISBN #1}\fi
\ifx \binits  \undefined \def \binits#1{#1}\fi
\ifx \bauthor  \undefined \def \bauthor#1{#1}\fi
\ifx \batitle  \undefined \def \batitle#1{#1}\fi
\ifx \bjtitle  \undefined \def \bjtitle#1{#1}\fi
\ifx \bvolume  \undefined \def \bvolume#1{\textbf{#1}}\fi
\ifx \byear  \undefined \def \byear#1{#1}\fi
\ifx \bissue  \undefined \def \bissue#1{#1}\fi
\ifx \bfpage  \undefined \def \bfpage#1{#1}\fi
\ifx \blpage  \undefined \def \blpage #1{#1}\fi
\ifx \burl  \undefined \def \burl#1{\textsf{#1}}\fi
\ifx \doiurl  \undefined \def \doiurl#1{\url{https://doi.org/#1}}\fi
\ifx \betal  \undefined \def \betal{\textit{et al.}}\fi
\ifx \binstitute  \undefined \def \binstitute#1{#1}\fi
\ifx \binstitutionaled  \undefined \def \binstitutionaled#1{#1}\fi
\ifx \bctitle  \undefined \def \bctitle#1{#1}\fi
\ifx \beditor  \undefined \def \beditor#1{#1}\fi
\ifx \bpublisher  \undefined \def \bpublisher#1{#1}\fi
\ifx \bbtitle  \undefined \def \bbtitle#1{#1}\fi
\ifx \bedition  \undefined \def \bedition#1{#1}\fi
\ifx \bseriesno  \undefined \def \bseriesno#1{#1}\fi
\ifx \blocation  \undefined \def \blocation#1{#1}\fi
\ifx \bsertitle  \undefined \def \bsertitle#1{#1}\fi
\ifx \bsnm \undefined \def \bsnm#1{#1}\fi
\ifx \bsuffix \undefined \def \bsuffix#1{#1}\fi
\ifx \bparticle \undefined \def \bparticle#1{#1}\fi
\ifx \barticle \undefined \def \barticle#1{#1}\fi
\bibcommenthead
\ifx \bconfdate \undefined \def \bconfdate #1{#1}\fi
\ifx \botherref \undefined \def \botherref #1{#1}\fi
\ifx \url \undefined \def \url#1{\textsf{#1}}\fi
\ifx \bchapter \undefined \def \bchapter#1{#1}\fi
\ifx \bbook \undefined \def \bbook#1{#1}\fi
\ifx \bcomment \undefined \def \bcomment#1{#1}\fi
\ifx \oauthor \undefined \def \oauthor#1{#1}\fi
\ifx \citeauthoryear \undefined \def \citeauthoryear#1{#1}\fi
\ifx \endbibitem  \undefined \def \endbibitem {}\fi
\ifx \bconflocation  \undefined \def \bconflocation#1{#1}\fi
\ifx \arxivurl  \undefined \def \arxivurl#1{\textsf{#1}}\fi
\csname PreBibitemsHook\endcsname

\bibitem{thomas2014tracheal}
\begin{barticle}
\bauthor{\bsnm{Thomas}, \binits{E.B.}},
\bauthor{\bsnm{Moss}, \binits{S.}}:
\batitle{Tracheal intubation}.
\bjtitle{Anaesth. Intensiv. Care Med.}
\bvolume{15}(\bissue{1}),
\bfpage{5}--\blpage{7}
(\byear{2014})
\end{barticle}
\endbibitem

\bibitem{caplan2003practice}
\begin{barticle}
\bauthor{\bsnm{Caplan}, \binits{R.A.}},
\bauthor{\bsnm{Benumof}, \binits{J.L.}},
\bauthor{\bsnm{Berry}, \binits{F.A.}},
\bauthor{\bsnm{Blitt}, \binits{C.D.}},
\bauthor{\bsnm{Bode}, \binits{R.H.}},
\bauthor{\bsnm{Cheney}, \binits{F.W.}},
\bauthor{\bsnm{Connis}, \binits{R.T.}},
\bauthor{\bsnm{Guidry}, \binits{O.F.}},
\bauthor{\bsnm{Nickinovich}, \binits{D.G.}},
\bauthor{\bsnm{Ovassapian}, \binits{A.}}:
\batitle{Practice guidelines for management of the difficult airway}.
\bjtitle{Anesthesiology}
\bvolume{98}(\bissue{1269-1277}),
\bfpage{2}
(\byear{2003})
\end{barticle}
\endbibitem

\bibitem{lu2021toward}
\begin{barticle}
\bauthor{\bsnm{Lu}, \binits{B.}},
\bauthor{\bsnm{Li}, \binits{B.}},
\bauthor{\bsnm{Chen}, \binits{W.}},
\bauthor{\bsnm{Jin}, \binits{Y.}},
\bauthor{\bsnm{Zhao}, \binits{Z.}},
\bauthor{\bsnm{Dou}, \binits{Q.}},
\bauthor{\bsnm{Heng}, \binits{P.-A.}},
\bauthor{\bsnm{Liu}, \binits{Y.}}:
\batitle{Toward image-guided automated suture grasping under complex
  environments: A learning-enabled and optimization-based holistic framework}.
\bjtitle{IEEE Transactions on Automation Science and Engineering}
\bvolume{19}(\bissue{4}),
\bfpage{3794}--\blpage{3808}
(\byear{2021})
\end{barticle}
\endbibitem

\bibitem{lai2021variable}
\begin{barticle}
\bauthor{\bsnm{Lai}, \binits{J.}},
\bauthor{\bsnm{Lu}, \binits{B.}},
\bauthor{\bsnm{Chu}, \binits{H.K.}}:
\batitle{Variable-stiffness control of a dual-segment soft robot using depth
  vision}.
\bjtitle{IEEE/ASME Transactions on Mechatronics}
\bvolume{27}(\bissue{2}),
\bfpage{1034}--\blpage{1045}
(\byear{2021})
\end{barticle}
\endbibitem

\bibitem{lu2022unified}
\begin{barticle}
\bauthor{\bsnm{Lu}, \binits{B.}},
\bauthor{\bsnm{Li}, \binits{B.}},
\bauthor{\bsnm{Dou}, \binits{Q.}},
\bauthor{\bsnm{Liu}, \binits{Y.}}:
\batitle{A unified monocular camera-based and pattern-free hand-to-eye
  calibration algorithm for surgical robots with rcm constraints}.
\bjtitle{IEEE/ASME Transactions on Mechatronics}
\bvolume{27}(\bissue{6}),
\bfpage{5124}--\blpage{5135}
(\byear{2022})
\end{barticle}
\endbibitem

\bibitem{bruce2022mmnet}
\begin{botherref}
\oauthor{\bsnm{Yu}, \binits{B.X.}},
\oauthor{\bsnm{Liu}, \binits{Y.}},
\oauthor{\bsnm{Zhang}, \binits{X.}},
\oauthor{\bsnm{Zhong}, \binits{S.-h.}},
\oauthor{\bsnm{Chan}, \binits{K.C.}}:
Mmnet: A model-based multimodal network for human action recognition in rgb-d
  videos.
IEEE Transactions on Pattern Analysis and Machine Intelligence
(2022)
\end{botherref}
\endbibitem

\bibitem{asgari2021deep}
\begin{barticle}
\bauthor{\bsnm{Asgari~Taghanaki}, \binits{S.}},
\bauthor{\bsnm{Abhishek}, \binits{K.}},
\bauthor{\bsnm{Cohen}, \binits{J.P.}},
\bauthor{\bsnm{Cohen-Adad}, \binits{J.}},
\bauthor{\bsnm{Hamarneh}, \binits{G.}}:
\batitle{Deep semantic segmentation of natural and medical images: a review}.
\bjtitle{Artif. Intell. Rev.}
\bvolume{54}(\bissue{1}),
\bfpage{137}--\blpage{178}
(\byear{2021})
\end{barticle}
\endbibitem

\bibitem{frangi2018simulation}
\begin{barticle}
\bauthor{\bsnm{Frangi}, \binits{A.F.}},
\bauthor{\bsnm{Tsaftaris}, \binits{S.A.}},
\bauthor{\bsnm{Prince}, \binits{J.L.}}:
\batitle{Simulation and synthesis in medical imaging}.
\bjtitle{IEEE Trans. Med. Image.}
\bvolume{37}(\bissue{3}),
\bfpage{673}--\blpage{679}
(\byear{2018})
\end{barticle}
\endbibitem

\bibitem{rehman2022organs}
\begin{barticle}
\bauthor{\bsnm{Rehman}, \binits{M.}},
\bauthor{\bsnm{Arsenault}, \binits{L.}},
\bauthor{\bsnm{Javan}, \binits{R.}}:
\batitle{Organs in color: utilizing free software and emerging multi jet fusion
  technology to color and surface label 3d-printed anatomical models}.
\bjtitle{J. Digit. Imaging}
\bvolume{35}(\bissue{6}),
\bfpage{1611}--\blpage{1622}
(\byear{2022})
\end{barticle}
\endbibitem

\bibitem{duriez2013control}
\begin{bchapter}
\bauthor{\bsnm{Duriez}, \binits{C.}}:
\bctitle{Control of elastic soft robots based on real-time finite element
  method}.
In: \bbtitle{Proc. IEEE Int. Conf. Robot. Autom. (ICRA)},
pp. \bfpage{3982}--\blpage{3987}
(\byear{2013})
\end{bchapter}
\endbibitem

\bibitem{zhao2020sim}
\begin{bchapter}
\bauthor{\bsnm{Zhao}, \binits{W.}},
\bauthor{\bsnm{Queralta}, \binits{J.P.}},
\bauthor{\bsnm{Westerlund}, \binits{T.}}:
\bctitle{Sim-to-real transfer in deep reinforcement learning for robotics: a
  survey}.
In: \bbtitle{Proc. IEEE Symp. Ser. Comput. Intell. (SSCI)},
pp. \bfpage{737}--\blpage{744}
(\byear{2020})
\end{bchapter}
\endbibitem

\bibitem{ganry2017use}
\begin{barticle}
\bauthor{\bsnm{Ganry}, \binits{L.}},
\bauthor{\bsnm{Hersant}, \binits{B.}},
\bauthor{\bsnm{Quilichini}, \binits{J.}},
\bauthor{\bsnm{Leyder}, \binits{P.}},
\bauthor{\bsnm{Meningaud}, \binits{J.}}:
\batitle{Use of the 3d surgical modelling technique with open-source software
  for mandibular fibula free flap reconstruction and its surgical guides}.
\bjtitle{J. Stomatol. Oral Maxillofac. Surg.}
\bvolume{118}(\bissue{3}),
\bfpage{197}--\blpage{202}
(\byear{2017})
\end{barticle}
\endbibitem

\bibitem{pierri2019bimaxillary}
\begin{barticle}
\bauthor{\bsnm{Pierri}, \binits{R.}},
\bauthor{\bsnm{Nogueira}, \binits{L.}},
\bauthor{\bsnm{Balan}, \binits{I.}},
\bauthor{\bsnm{Iwaki}, \binits{L.}}, \betal:
\batitle{Bimaxillary orthognatic surgery planned with the software blender,
  through the addon ortogonblender}.
\bjtitle{Int. J. Oral Maxillofac. Surg.}
\bvolume{48},
\bfpage{254}
(\byear{2019})
\end{barticle}
\endbibitem

\bibitem{chen2021understanding}
\begin{botherref}
\oauthor{\bsnm{Chen}, \binits{X.}},
\oauthor{\bsnm{Hu}, \binits{J.}},
\oauthor{\bsnm{Jin}, \binits{C.}},
\oauthor{\bsnm{Li}, \binits{L.}},
\oauthor{\bsnm{Wang}, \binits{L.}}:
Understanding domain randomization for sim-to-real transfer.
arXiv preprint arXiv:2110.03239
(2021)
\end{botherref}
\endbibitem

\bibitem{tobin2017domain}
\begin{bchapter}
\bauthor{\bsnm{Tobin}, \binits{J.}},
\bauthor{\bsnm{Fong}, \binits{R.}},
\bauthor{\bsnm{Ray}, \binits{A.}},
\bauthor{\bsnm{Schneider}, \binits{J.}},
\bauthor{\bsnm{Zaremba}, \binits{W.}},
\bauthor{\bsnm{Abbeel}, \binits{P.}}:
\bctitle{Domain randomization for transferring deep neural networks from
  simulation to the real world}.
In: \bbtitle{IEEE/RSJ Int. Conf. Intell. Robot. Syst. (IROS)},
pp. \bfpage{23}--\blpage{30}
(\byear{2017}).
\bcomment{IEEE}
\end{bchapter}
\endbibitem

\bibitem{yang2020fda}
\begin{bchapter}
\bauthor{\bsnm{Yang}, \binits{Y.}},
\bauthor{\bsnm{Soatto}, \binits{S.}}:
\bctitle{Fda: Fourier domain adaptation for semantic segmentation}.
In: \bbtitle{Proc. IEEE Comput. Soc. Conf. Comput. Vis. Pattern Recognit.
  (CVPR)},
pp. \bfpage{4085}--\blpage{4095}
(\byear{2020})
\end{bchapter}
\endbibitem

\bibitem{geng2011daml}
\begin{barticle}
\bauthor{\bsnm{Geng}, \binits{B.}},
\bauthor{\bsnm{Tao}, \binits{D.}},
\bauthor{\bsnm{Xu}, \binits{C.}}:
\batitle{Daml: Domain adaptation metric learning}.
\bjtitle{IEEE Trans. Image Process.}
\bvolume{20}(\bissue{10}),
\bfpage{2980}--\blpage{2989}
(\byear{2011})
\end{barticle}
\endbibitem

\bibitem{long2015learning}
\begin{bchapter}
\bauthor{\bsnm{Long}, \binits{M.}},
\bauthor{\bsnm{Cao}, \binits{Y.}},
\bauthor{\bsnm{Wang}, \binits{J.}},
\bauthor{\bsnm{Jordan}, \binits{M.}}:
\bctitle{Learning transferable features with deep adaptation networks}.
In: \bbtitle{Proc. Int. Conf. Mach. Learn. (ICML)},
pp. \bfpage{97}--\blpage{105}
(\byear{2015}).
\bcomment{PMLR}
\end{bchapter}
\endbibitem

\bibitem{zellinger2017central}
\begin{botherref}
\oauthor{\bsnm{Zellinger}, \binits{W.}},
\oauthor{\bsnm{Grubinger}, \binits{T.}},
\oauthor{\bsnm{Lughofer}, \binits{E.}},
\oauthor{\bsnm{Natschl{\"a}ger}, \binits{T.}},
\oauthor{\bsnm{Saminger-Platz}, \binits{S.}}:
Central moment discrepancy (cmd) for domain-invariant representation learning.
arXiv preprint arXiv:1702.08811
(2017)
\end{botherref}
\endbibitem

\bibitem{zou2018unsupervised}
\begin{bchapter}
\bauthor{\bsnm{Zou}, \binits{Y.}},
\bauthor{\bsnm{Yu}, \binits{Z.}},
\bauthor{\bsnm{Kumar}, \binits{B.}},
\bauthor{\bsnm{Wang}, \binits{J.}}:
\bctitle{Unsupervised domain adaptation for semantic segmentation via
  class-balanced self-training}.
In: \bbtitle{Proc. Eur. Conf. Comput. Vis. (ECCV)},
pp. \bfpage{289}--\blpage{305}
(\byear{2018})
\end{bchapter}
\endbibitem

\bibitem{wu2018dcan}
\begin{bchapter}
\bauthor{\bsnm{Wu}, \binits{Z.}},
\bauthor{\bsnm{Han}, \binits{X.}},
\bauthor{\bsnm{Lin}, \binits{Y.-L.}},
\bauthor{\bsnm{Uzunbas}, \binits{M.G.}},
\bauthor{\bsnm{Goldstein}, \binits{T.}},
\bauthor{\bsnm{Lim}, \binits{S.N.}},
\bauthor{\bsnm{Davis}, \binits{L.S.}}:
\bctitle{Dcan: Dual channel-wise alignment networks for unsupervised scene
  adaptation}.
In: \bbtitle{Proc. Eur. Conf. Comput. Vis. (ECCV)},
pp. \bfpage{518}--\blpage{534}
(\byear{2018})
\end{bchapter}
\endbibitem

\bibitem{sankaranarayanan2018learning}
\begin{bchapter}
\bauthor{\bsnm{Sankaranarayanan}, \binits{S.}},
\bauthor{\bsnm{Balaji}, \binits{Y.}},
\bauthor{\bsnm{Jain}, \binits{A.}},
\bauthor{\bsnm{Lim}, \binits{S.N.}},
\bauthor{\bsnm{Chellappa}, \binits{R.}}:
\bctitle{Learning from synthetic data: Addressing domain shift for semantic
  segmentation}.
In: \bbtitle{Proc. IEEE Comput. Soc. Conf. Comput. Vis. Pattern Recognit.
  (CVPR)},
pp. \bfpage{3752}--\blpage{3761}
(\byear{2018})
\end{bchapter}
\endbibitem

\bibitem{vu2019advent}
\begin{bchapter}
\bauthor{\bsnm{Vu}, \binits{T.-H.}},
\bauthor{\bsnm{Jain}, \binits{H.}},
\bauthor{\bsnm{Bucher}, \binits{M.}},
\bauthor{\bsnm{Cord}, \binits{M.}},
\bauthor{\bsnm{P{\'e}rez}, \binits{P.}}:
\bctitle{Advent: Adversarial entropy minimization for domain adaptation in
  semantic segmentation}.
In: \bbtitle{Proc. IEEE Comput. Soc. Conf. Comput. Vis. Pattern Recognit.
  (CVPR)},
pp. \bfpage{2517}--\blpage{2526}
(\byear{2019})
\end{bchapter}
\endbibitem

\bibitem{hoffman2018cycada}
\begin{bchapter}
\bauthor{\bsnm{Hoffman}, \binits{J.}},
\bauthor{\bsnm{Tzeng}, \binits{E.}},
\bauthor{\bsnm{Park}, \binits{T.}},
\bauthor{\bsnm{Zhu}, \binits{J.-Y.}},
\bauthor{\bsnm{Isola}, \binits{P.}},
\bauthor{\bsnm{Saenko}, \binits{K.}},
\bauthor{\bsnm{Efros}, \binits{A.}},
\bauthor{\bsnm{Darrell}, \binits{T.}}:
\bctitle{Cycada: Cycle-consistent adversarial domain adaptation}.
In: \bbtitle{Proc. Int. Conf. Mach. Learn. (ICML)},
pp. \bfpage{1989}--\blpage{1998}
(\byear{2018}).
\bcomment{Pmlr}
\end{bchapter}
\endbibitem

\bibitem{richter2016playing}
\begin{bchapter}
\bauthor{\bsnm{Richter}, \binits{S.R.}},
\bauthor{\bsnm{Vineet}, \binits{V.}},
\bauthor{\bsnm{Roth}, \binits{S.}},
\bauthor{\bsnm{Koltun}, \binits{V.}}:
\bctitle{Playing for data: Ground truth from computer games}.
In: \bbtitle{Proc. Eur. Conf. Comput. Vis. (ECCV)},
pp. \bfpage{102}--\blpage{118}
(\byear{2016}).
\bcomment{Springer}
\end{bchapter}
\endbibitem

\bibitem{ros2016synthia}
\begin{bchapter}
\bauthor{\bsnm{Ros}, \binits{G.}},
\bauthor{\bsnm{Sellart}, \binits{L.}},
\bauthor{\bsnm{Materzynska}, \binits{J.}},
\bauthor{\bsnm{Vazquez}, \binits{D.}},
\bauthor{\bsnm{Lopez}, \binits{A.M.}}:
\bctitle{The synthia dataset: A large collection of synthetic images for
  semantic segmentation of urban scenes}.
In: \bbtitle{Proc. IEEE Comput. Soc. Conf. Comput. Vis. Pattern Recognit.
  (CVPR)},
pp. \bfpage{3234}--\blpage{3243}
(\byear{2016})
\end{bchapter}
\endbibitem

\bibitem{cordts2016cityscapes}
\begin{bchapter}
\bauthor{\bsnm{Cordts}, \binits{M.}},
\bauthor{\bsnm{Omran}, \binits{M.}},
\bauthor{\bsnm{Ramos}, \binits{S.}},
\bauthor{\bsnm{Rehfeld}, \binits{T.}},
\bauthor{\bsnm{Enzweiler}, \binits{M.}},
\bauthor{\bsnm{Benenson}, \binits{R.}},
\bauthor{\bsnm{Franke}, \binits{U.}},
\bauthor{\bsnm{Roth}, \binits{S.}},
\bauthor{\bsnm{Schiele}, \binits{B.}}:
\bctitle{The cityscapes dataset for semantic urban scene understanding}.
In: \bbtitle{Proc. IEEE Comput. Soc. Conf. Comput. Vis. Pattern Recognit.
  (CVPR)},
pp. \bfpage{3213}--\blpage{3223}
(\byear{2016})
\end{bchapter}
\endbibitem

\bibitem{li2019bidirectional}
\begin{bchapter}
\bauthor{\bsnm{Li}, \binits{Y.}},
\bauthor{\bsnm{Yuan}, \binits{L.}},
\bauthor{\bsnm{Vasconcelos}, \binits{N.}}:
\bctitle{Bidirectional learning for domain adaptation of semantic
  segmentation}.
In: \bbtitle{Proc. IEEE Comput. Soc. Conf. Comput. Vis. Pattern Recognit.
  (CVPR)},
pp. \bfpage{6936}--\blpage{6945}
(\byear{2019})
\end{bchapter}
\endbibitem

\bibitem{zhu2005semi}
\begin{botherref}
\oauthor{\bsnm{Zhu}, \binits{X.J.}}:
Semi-supervised learning literature survey.
Technical report, University of Wisconsin-Madison Department of Computer
  Sciences
(2005)
\end{botherref}
\endbibitem

\bibitem{springenberg2015unsupervised}
\begin{botherref}
\oauthor{\bsnm{Springenberg}, \binits{J.T.}}:
Unsupervised and semi-supervised learning with categorical generative
  adversarial networks.
arXiv preprint arXiv:1511.06390
(2015)
\end{botherref}
\endbibitem

\bibitem{an2021artflow}
\begin{bchapter}
\bauthor{\bsnm{An}, \binits{J.}},
\bauthor{\bsnm{Huang}, \binits{S.}},
\bauthor{\bsnm{Song}, \binits{Y.}},
\bauthor{\bsnm{Dou}, \binits{D.}},
\bauthor{\bsnm{Liu}, \binits{W.}},
\bauthor{\bsnm{Luo}, \binits{J.}}:
\bctitle{Artflow: Unbiased image style transfer via reversible neural flows}.
In: \bbtitle{Proc. IEEE Comput. Soc. Conf. Comput. Vis. Pattern Recognit.
  (CVPR)},
pp. \bfpage{862}--\blpage{871}
(\byear{2021})
\end{bchapter}
\endbibitem

\bibitem{unpublished}
\begin{botherref}
\oauthor{\bsnm{Lai}, \binits{J.}},
\oauthor{\bsnm{Ren}, \binits{T.-A.}},
\oauthor{\bsnm{Yue}, \binits{W.}},
\oauthor{\bsnm{Su}, \binits{S.}},
\oauthor{\bsnm{Chan}, \binits{J.Y.K.}},
\oauthor{\bsnm{Ren}, \binits{H.}}:
Sim-to-real transfer of soft robotic navigation strategies that learns from the
  virtual eye-in-hand vision.
Unpublished
(2023)
\end{botherref}
\endbibitem

\bibitem{lai2022constrained}
\begin{barticle}
\bauthor{\bsnm{Lai}, \binits{J.}},
\bauthor{\bsnm{Lu}, \binits{B.}},
\bauthor{\bsnm{Zhao}, \binits{Q.}},
\bauthor{\bsnm{Chu}, \binits{H.K.}}:
\batitle{Constrained motion planning of a cable-driven soft robot with
  compressible curvature modeling}.
\bjtitle{IEEE Robot. Autom. Lett.}
\bvolume{7}(\bissue{2}),
\bfpage{4813}--\blpage{4820}
(\byear{2022})
\end{barticle}
\endbibitem

\bibitem{allan2019endovis17}
\begin{botherref}
\oauthor{\bsnm{Allan}, \binits{M.}},
\oauthor{\bsnm{Shvets}, \binits{A.}},
\oauthor{\bsnm{Kurmann}, \binits{T.}},
\oauthor{\bsnm{Zhang}, \binits{Z.}},
\oauthor{\bsnm{Duggal}, \binits{R.}},
\oauthor{\bsnm{Su}, \binits{Y.-H.}},
\oauthor{\bsnm{Rieke}, \binits{N.}},
\oauthor{\bsnm{Laina}, \binits{I.}},
\oauthor{\bsnm{Kalavakonda}, \binits{N.}},
\oauthor{\bsnm{Bodenstedt}, \binits{S.}}, et al.:
2017 robotic instrument segmentation challenge.
arXiv preprint arXiv:1902.06426
(2019)
\end{botherref}
\endbibitem

\bibitem{allan2020endovis18}
\begin{botherref}
\oauthor{\bsnm{Allan}, \binits{M.}},
\oauthor{\bsnm{Kondo}, \binits{S.}},
\oauthor{\bsnm{Bodenstedt}, \binits{S.}},
\oauthor{\bsnm{Leger}, \binits{S.}},
\oauthor{\bsnm{Kadkhodamohammadi}, \binits{R.}},
\oauthor{\bsnm{Luengo}, \binits{I.}},
\oauthor{\bsnm{Fuentes}, \binits{F.}},
\oauthor{\bsnm{Flouty}, \binits{E.}},
\oauthor{\bsnm{Mohammed}, \binits{A.}},
\oauthor{\bsnm{Pedersen}, \binits{M.}}, et al.:
2018 robotic scene segmentation challenge.
arXiv preprint arXiv:2001.11190
(2020)
\end{botherref}
\endbibitem

\bibitem{3dmodel}
\begin{botherref}
\oauthor{\bsnm{{University of Dundee, School of Medicine}}}:
{Pharynx and Floor of Mouth}.
\url{https://skfb.ly/6QXqr}.
Accessed: 2022-08-01
\end{botherref}
\endbibitem

\bibitem{ghiasi2017exploring}
\begin{botherref}
\oauthor{\bsnm{Ghiasi}, \binits{G.}},
\oauthor{\bsnm{Lee}, \binits{H.}},
\oauthor{\bsnm{Kudlur}, \binits{M.}},
\oauthor{\bsnm{Dumoulin}, \binits{V.}},
\oauthor{\bsnm{Shlens}, \binits{J.}}:
Exploring the structure of a real-time, arbitrary neural artistic stylization
  network.
arXiv preprint arXiv:1705.06830
(2017)
\end{botherref}
\endbibitem

\bibitem{li2017universal}
\begin{botherref}
\oauthor{\bsnm{Li}, \binits{Y.}},
\oauthor{\bsnm{Fang}, \binits{C.}},
\oauthor{\bsnm{Yang}, \binits{J.}},
\oauthor{\bsnm{Wang}, \binits{Z.}},
\oauthor{\bsnm{Lu}, \binits{X.}},
\oauthor{\bsnm{Yang}, \binits{M.-H.}}:
Universal style transfer via feature transforms.
Adv. Neural Info. Processing Syst.
\textbf{30}
(2017)
\end{botherref}
\endbibitem

\bibitem{huang2017arbitrary}
\begin{bchapter}
\bauthor{\bsnm{Huang}, \binits{X.}},
\bauthor{\bsnm{Belongie}, \binits{S.}}:
\bctitle{Arbitrary style transfer in real-time with adaptive instance
  normalization}.
In: \bbtitle{Proc. IEEE Int. Conf. Compt. Vis. (ICCV)},
pp. \bfpage{1501}--\blpage{1510}
(\byear{2017})
\end{bchapter}
\endbibitem

\bibitem{liao2017visual}
\begin{botherref}
\oauthor{\bsnm{Liao}, \binits{J.}},
\oauthor{\bsnm{Yao}, \binits{Y.}},
\oauthor{\bsnm{Yuan}, \binits{L.}},
\oauthor{\bsnm{Hua}, \binits{G.}},
\oauthor{\bsnm{Kang}, \binits{S.B.}}:
Visual attribute transfer through deep image analogy.
arXiv preprint arXiv:1705.01088
(2017)
\end{botherref}
\endbibitem

\bibitem{kingma2018glow}
\begin{botherref}
\oauthor{\bsnm{Kingma}, \binits{D.P.}},
\oauthor{\bsnm{Dhariwal}, \binits{P.}}:
Glow: Generative flow with invertible 1x1 convolutions.
Adv. Neural Info. Processing Syst.
\textbf{31}
(2018)
\end{botherref}
\endbibitem

\bibitem{dinh2014nice}
\begin{botherref}
\oauthor{\bsnm{Dinh}, \binits{L.}},
\oauthor{\bsnm{Krueger}, \binits{D.}},
\oauthor{\bsnm{Bengio}, \binits{Y.}}:
Nice: Non-linear independent components estimation.
arXiv preprint arXiv:1410.8516
(2014)
\end{botherref}
\endbibitem

\bibitem{chen2017deeplab}
\begin{barticle}
\bauthor{\bsnm{Chen}, \binits{L.-C.}},
\bauthor{\bsnm{Papandreou}, \binits{G.}},
\bauthor{\bsnm{Kokkinos}, \binits{I.}},
\bauthor{\bsnm{Murphy}, \binits{K.}},
\bauthor{\bsnm{Yuille}, \binits{A.L.}}:
\batitle{Deeplab: Semantic image segmentation with deep convolutional nets,
  atrous convolution, and fully connected crfs}.
\bjtitle{IEEE Trans. Pattern Anal. Mach. Intell.}
\bvolume{40}(\bissue{4}),
\bfpage{834}--\blpage{848}
(\byear{2017})
\end{barticle}
\endbibitem

\bibitem{he2016deep}
\begin{bchapter}
\bauthor{\bsnm{He}, \binits{K.}},
\bauthor{\bsnm{Zhang}, \binits{X.}},
\bauthor{\bsnm{Ren}, \binits{S.}},
\bauthor{\bsnm{Sun}, \binits{J.}}:
\bctitle{Deep residual learning for image recognition}.
In: \bbtitle{Proc. IEEE Comput. Soc. Conf. Comput. Vis. Pattern Recognit.
  (CVPR)},
pp. \bfpage{770}--\blpage{778}
(\byear{2016})
\end{bchapter}
\endbibitem

\bibitem{kingma2014adam}
\begin{botherref}
\oauthor{\bsnm{Kingma}, \binits{D.P.}},
\oauthor{\bsnm{Ba}, \binits{J.}}:
Adam: A method for stochastic optimization.
arXiv preprint arXiv:1412.6980
(2014)
\end{botherref}
\endbibitem

\end{thebibliography}

\bigskip

\end{document}